\documentclass[letterpaper, 10 pt, conference]{ieeeconf}  

\IEEEoverridecommandlockouts                              

\usepackage{amsmath} 
\usepackage{amssymb}  
\usepackage{booktabs}
\usepackage{graphicx}
\usepackage{multirow}
\usepackage[table,xcdraw]{xcolor}
\usepackage{cite}

\title{\LARGE \bf
Generalizability Analysis of Graph-based Trajectory Predictor with Vectorized Representation
}

\author{Juanwu Lu$^{1}$, Wei Zhan$^{2}$, Masayoshi Tomizuka$^{2}$, and Yeping Hu$^{2}$
\thanks{$^{1}$ J. Lu is with the Department of Civil and Environmental Engineering, University of California, Berkeley, CA 94720 USA.
        {[\tt\small juanwu\_lu@berkeley.edu]}}%
\thanks{$^{2}$ M. Tomizuka, W. Zhan, and Y. Hu are with the Department of Mechanical Engineering, University of California, Berkeley, CA 94720 USA.
        {[\tt\small wzhan, tomizuka, yeping\_hu@berkeley.edu]}}%
}

\begin{document}

\maketitle
\thispagestyle{empty}
\pagestyle{empty}

\begin{abstract}

Trajectory prediction is one of the essential tasks for autonomous vehicles. Recent progress in machine learning gave birth to a series of advanced trajectory prediction algorithms. Lately, the effectiveness of using graph neural networks (GNNs) with vectorized representations for trajectory prediction has been demonstrated by many researchers. Nonetheless, these algorithms either pay little attention to models' generalizability across various scenarios or simply assume training and test data follow similar statistics. In fact, when test scenarios are unseen or Out-of-Distribution (OOD), the resulting train-test domain shift usually leads to significant degradation in prediction performance, which will impact downstream modules and eventually lead to severe accidents. Therefore, it is of great importance to thoroughly investigate the prediction models in terms of their generalizability, which can not only help identify their weaknesses but also provide insights on how to improve these models. This paper proposes a generalizability analysis framework using feature attribution methods to help interpret black-box models. For the case study, we provide an in-depth generalizability analysis of one of the state-of-the-art graph-based trajectory predictors that utilize vectorized representation. Results show significant performance degradation due to domain shift, and feature attribution provides insights to identify potential causes of these problems. Finally, we conclude the common prediction challenges and how weighting biases induced by the training process can deteriorate the accuracy.

\end{abstract}

\section{INTRODUCTION}

Trajectory prediction refers to inferring the future movements of surrounding vehicles based on previous observations and possible interactions, which is critical for the safety and reliability of autonomous vehicles. By its paradigm, existing trajectory prediction methods include rule-based and data-driven models. Although the former usually combines physical models with maneuver intention recognition to improve performance \cite{6696982}, the long-term prediction may diverge or even worse than constant velocity inference \cite{andersson2018predicting}.

In the past decade, substantial progress in machine learning has promoted the applications of data-driven models. Early studies \cite{Alahi_2016_CVPR,bojarski2017explaining,8793868} adopt Convolutional Neural Networks (CNNs) for graphical feature extraction and Recurrent Neural Networks (RNNs) for sequential trajectory encoding and decoding processes. To further improve efficiency, \cite{Djuric_2020_WACV} used CNN to encode an RGB representation of the scene context, concatenated it with the trajectory embedding, and fed them into a simple feed-forward network to derive the prediction. Nonetheless, the limitation of using such rasterized representations and CNN architectures is that the approach can only effectively extract connections among neighboring pixels and struggles with broader global interactions among vehicles and the context items. The latest methods use graph neural networks (GNNs) and attention mechanisms to model higher-level interactions. Vectorized representations of the trajectories and the scenario contexts emerge as they help better capture the structural information. These models describe interaction within context as learnable graphs and use data-driven message passing to infer node-, edge-, and graph-level representations from datasets.



Despite the promising prediction performance, most existing models are based on a fundamental assumption that training and test data are \emph{Identically and Independently Distributed} (IID). For example, the training and testing cases can have exact or similar driving locations and traffic conditions, which may not be the case in the real world. Consistently enlarging the dataset and promoting scenario diversity can, to some extent, mitigate the issue, but data imbalance and the long tail problem are still inevitable. Other works highlighted transferable and adaptable feature engineering and model designs \cite{https://doi.org/10.48550/arxiv.2004.03053, https://doi.org/10.48550/arxiv.2202.05140}. Nevertheless, this area is missing a complete framework for thorough generalizability analysis and understanding of the limitations. 

Therefore, this paper proposes an analytical framework that combines cross-scenario validation with feature attribution techniques. A model's generalizability is cross-validated by training and testing across various traffic scenarios. Further analysis of potential causes of generalization bottlenecks is achieved through a proposed generalized version of the integrated gradients (IG) method for trajectory prediction, which directly measures how a model attributes its prediction accuracy to the input features. We apply this framework to analyze VectorNet, a state-of-the-art graph-based trajectory prediction model with vectorized representations and observe a significant performance degradation. Quantitative and qualitative analysis of feature attributions reveals potential causes that lead to generalization bottlenecks.

\begin{figure*}[t]
    \centering
    \includegraphics[width=6.8in]{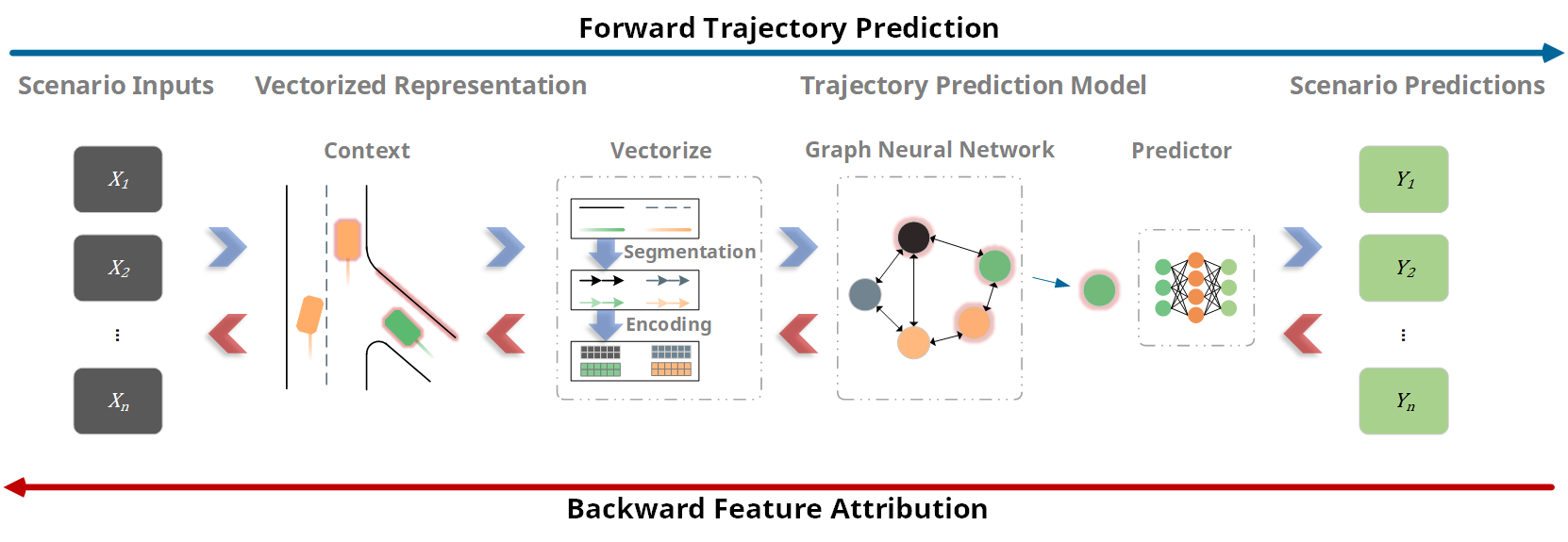}
    \caption{Illustration of the analysis framework. Objects in a traffic scenario are represented as a set of vectors through segmentation and encoding. The forward trajectory prediction takes these vectors as inputs to the model and derives results. The backward feature attribution estimates how the model attributes its prediction accuracy to the input vectors based on the predictions.}
    \label{fig:framework}
\end{figure*}

\section{RELATED WORKS}

\subsection{Graph Neural Networks for Trajectory Prediction}

The latest trajectory prediction methods utilize graph neural networks (GNNs) to tackle the limitations of the traditional CNNs and RNNs. These methods often use vectorized encoding to represent spatial phenomena. Compared to raster data, vectors can more precisely describe lines and edges and can capture meaningful topological connections \cite{maffini1987raster}. Such vectorized representation helps to leverage the ability of GNNs to learn graph structures among items. VectorNet \cite{Gao_2020_CVPR} is one of the earliest methods to use GNNs with vectorized representation. It introduced a hierarchical network architecture consisting of subgraphs and a global attention graph to learn different levels of interactions. In serveral other studies, one can find similar use cases of combining GNNs with vector representations. Li et al. \cite{li2020evolvegraph} proposed using a directed, fully-connected observation graph and an interaction graph to encode vectorized information and interactions in the scenario. Yu et al. \cite{10.1007/978-3-030-58610-2_30} proposed the STAR framework that uses the Transformer attention mechanism and graph convolutional layers to separately model the spatial and temporal interactions of agents and road segment vectors. Huang et al.\cite{Huang_2019_ICCV} used fixed-length vectors with relative position embeddings and proposed a Spatio-Temporal Graph Attention network based on sequence-to-sequence architecture.

\subsection{Generalizabililty Studies on GNNs}

GNNs can suffer substantial performance drops when distribution shifts appear between training and testing datasets \cite{wu2018moleculenet, NEURIPS2020_fb60d411, pmlr-v139-koh21a}. Xu et al. \cite{xu2019powerful} evaluate various simple graph structures on their representational power and generalization ability quantified by test set performance. Du et al. \cite{NEURIPS2019_663fd3c5} proposed a new class of graph kernel to address generalizability and test it on node-level classification tasks. However, both assume training and testing data are IID samples from a latent distribution. Other works explain the degradation further by looking into the correlations between encoded features and predicted labels. Some identify spurious correlations between irrelevant representations and labels as the primary cause \cite{cogswell2016reducing}, while others attribute it to disentangled graph representation learning \cite{6472238}. Ma et al. \cite{ma2021subgroup} recently established a novel PAC-Bayesian analysis under a non-IID semi-supervised learning setup to derive a generalization bound for node-level tasks using GNNs. Li et al. \cite{li2021oodgnn} combined nonlinear decorrelation with a global weight estimator to improve the generalization ability of graph networks. Despite emerging interests in the problem, most existing works theoretically understand this issue in a classification context, while attention on trajectory prediction is relatively limited.

\subsection{Feature Attribution Methods}

Lack of generalizability is a challenge to real-world applications, and explaining how a model reacts to input features helps practitioners fix this issue \cite{cadamuro2016debugging}. Existing feature attribution methods can be either decomposition- or gradient-based. The former usually uses propagation algorithms to approximate the model by combining a set of linear functions; examples are Guided Backpropagation \cite{springenberg2015striving}, Layer-wise relevance propagation \cite{Nam_Gur_Choi_Wolf_Lee_2020}, etc. However, these methods could be insensitive to the inputs or the model parameters \cite{adebayo2020sanity}. On the contrary, gradient-based methods such as the Grad-CAM proposed by Selvaraju et al. directly utilize the gradients of the output to the inputs as the attributions \cite{GradCAM2019}. Smilkov et al. introduced SmoothGrad \cite{smilkov2017smoothgrad}, which reduces noise by averaging the gradients calculated with Gaussian noises. But these methods fail to satisfy some axiomatic properties when strong nonlinearity and gradient saturation exist \cite{sundararajan2016gradients}.

The latest research manages to unify feature attribution methods from the perspective of Shapley value in game theory\cite{lundberg2017unified, pmlr-v70-sundararajan17a}. The idea is to infer the share of importance given each feature with respect to the output. Existing works utilize these methods for interpreting and examining image classification \cite{kokhlikyan2020captum} and natural language processing \cite{sanyal2021discretized, nayak2021using} algorithms. However, most of these studies discuss the implementation in a classification context, which is essentially different from the formulation of a trajectory prediction problem (see Section III.A). In this paper, we generalize an integrated gradient method specifically for trajectory prediction tasks.

\begin{table*}[t]
    \centering
    \caption{Training and Test Dataset}
    \resizebox{\textwidth}{!}{%
    \begin{tabular}{@{}lcccccccc@{}}
    \toprule
    \multicolumn{1}{c}{}                                         &                                                   &                                          & \multicolumn{3}{c}{\textbf{Training Set}}                 & \multicolumn{3}{c}{\textbf{Test Set}}                     \\ \cmidrule(l){4-9} 
    \multicolumn{1}{c}{\multirow{-2}{*}{\textbf{Scenario Name}}} & \multirow{-2}{*}{\textbf{Type}}                   & \multirow{-2}{*}{\textbf{Map Polylines}} & \textbf{Size} & \textbf{Vehicles} & \textbf{Pedestrians and Bicycles} & \textbf{Size} & \textbf{Vehicles} & \textbf{Pedestrians and Bicycles} \\ \midrule
    \textbf{CHN Merging ZS0}                                     & \cellcolor[HTML]{E2F2D5}Expressway (On-ramp)      & 71                                       & 76439         & 78241             & 0                     & 20402         & 20782             & 0                     \\
    \textbf{CHN Merging ZS2}                                     & \cellcolor[HTML]{E2F2D5}Expressway (Mainline)     & 71                                       & 45618         & 48181             & 0                     & 19384         & 19997             & 0                     \\
    \textbf{USA Intersection GL}                                 & \cellcolor[HTML]{F6C2C2}Unsignalized Intersection & 163                                      & 127567        & 135459            & 3879                  & 30698         & 32647             & 853                   \\
    \textbf{USA Intersection MA}                                 & \cellcolor[HTML]{F6C2C2}Unsignalized Intersection & 139                                      & 48784         & 51084             & 1733                  & 9246          & 9734              & 430                   \\
    \textbf{USA Roundabout FT}                                   & \cellcolor[HTML]{F9FBBA}Roundabout                & 145                                      & 99964         & 105518            & 5201                  & 25893         & 27301             & 1216                  \\
    \textbf{DEU Roundabout OF}                                   & \cellcolor[HTML]{F9FBBA}Roundabout                & 113                                      & 10839         & 11630             & 666                   & 1999          & 2161              & 87                    \\ \bottomrule
    \end{tabular}
    }
    \label{tab:1}
\end{table*}

\section{METHODS}


This section first provides a general definition of the trajectory prediction problem with graph-based models and vectorized representation. We then simplify our discussion by selecting a representative graph neural network model for our implementation. Finally, we generalize the definition of feature attribution methods to the context of trajectory prediction and briefly introduce the method we used in our framework.

\subsection{Trajectory Prediction with Graph-based Models and Vectorized Representations}

\textbf{Problem Statement.} Given an observation horizon $T_h$ and a prediction horizon $T_f$, a trajectory prediction problem is a kind of regression problem in which the future locations of individuals $Y_{T_h+1:T_f}=\{y_{k, T_h+1:T_f}|k\in S\}$ in a traffic scenario $S$ are deduced based on a feature matrix of previous observations $X_{0:T_h}$. The goal is to fit a model $f$ that maps observation to predictions with high accuracy:
\begin{equation}
    f^*=\underset{f^\prime\in\mathcal{F}}{\text{argmin}}\mathcal{L}(Y_{T_h+1:T_f}, f^\prime(X_{0:T_h})),
\end{equation}
where $\mathcal{L}$ is a measure of prediction error. Generally, any traffic scenario can be described using a set of polyline chains $P=\{p_1,p_2,\ldots,p_n\}$ such as roadway markings, traffic signs, and vehicle trajectories. A polyline object can further split into a set of vectors $p_i=\{v_1^0,v_2^0,\ldots,v_m^0\}$, each as a segment of the original polyline. Following this pattern, vectors can also split into sub-vectors until satisfying some preset granularity. We refer to this process as the segmentation procedure to derive vectorized representation (see Fig. 1).

Suppose a polyline goes through recursive segmentation $H-1$ times, creating an $H$-level hierarchy. Items in the lowest level are vectors satisfying the granularity. We encode them as one-dimensional column vectors $x_j^H\in\mathbb{R}^L$. Each vector must include the origin and destination coordinates and can optionally hold information such as types, dynamics, etc. For each level up, a vector in the current level can be viewed as a subgraph $x_j^{h}=\mathcal{G}^h(X_j^{h+1}, A_j^{h})$, where $X_j^{h+1}$ is the node feature matrix of sub-vectors, and $A_j^h$ is the adjacency matrix. Finally, the top-level graph consists of a feature matrix $X^0\in\mathbb{R}^{|P|\times L_0}$ of all polylines and an adjacency matrix $A^0$ denoting mutual interactions.

The graph-based trajectory prediction model has a graph neural network $g$ that learns the optimal combinations of feature matrices in a bottom-up procedure. In each hierarchy, two types of transformation functions are learned in a data-driven paradigm: encoding functions $\phi_{enc}$ and aggregation functions $\phi_{agg}$. The encoding functions transform node feature matrices into hidden spaces. The aggregation functions take node features and adjacency matrix to generate a one-dimensional column vector representing a subgraph.
\begin{equation}
    X_j^h\leftarrow \phi_{enc}(X_j^h).
\end{equation}
\begin{equation}
    x_j^h\leftarrow \phi_{agg}(X_j^{h+1}, A_j^{h}).
\end{equation}

The predictor model uses learned node features in the top-level graph to infer future trajectories. If we denote the input feature hierarchy by $X$ and the initial adjacency matrix by $A$, The predictor model is given by:
\begin{equation}
    \hat{Y}_{T_h+1:T_f}=h(g(X_{0:T_h}, A)).
\end{equation}

\textbf{Implementation of VectorNet.} This paper focuses on a state-of-the-art hierarchical graph neural network, VectorNet, and examines its generalizability. Specifically, traffic scenarios consist of context polylines $c^{(j)}=\{c_n^{(j)}\}_{n=1}^N$ (e.g., border markings, lane markings, etc.) and trajectory polylines $r^{j}=\{r_t^{j}\}_{t=1}^{T_h}$. Here, $N$ and $T_h$ are the numbers of segments and historical frames, respectively. Polylines go through segmentation once and are represented by sets of vectors:
\begin{equation}
c_i^{(j)}=\left[\mathbf{x}_i^o,\mathbf{x}_i^d,\mathbf{a}^{(j)},\mathbf{0},j\right],\quad r_i^{(j)}=\left[\mathbf{x}_i^o,\mathbf{x}_i^d,\mathbf{a}^{(j)},\mathbf{s}_i,j\right],
\end{equation}
Vectors in the above polylines share a similar structure. The first four elements of a node are Cartesian coordinate pairs $\mathbf{x}_i^o$ and $\mathbf{x}_i^d$ related to the vector head and tail. A one-hot vector $\mathbf{a}^{(j)}$ is then used to identify a specific polyline type. Trajectory vectors have a state array $\mathbf{s}_i$ consisting of motion status such as current velocity, heading, and vehicle size, where context vectors have zero padding in the corresponding positions. The last element of a vector is its integer polyline id.

The vector level transformation uses two separate subgraphs to encode latent representations for the two types of polylines. Each subgraph is a stack of consecutive layers consisting of a Multi-Layer Perceptron (MLP) encoder $g_{enc}$, max-pooling aggregation $\varphi_{agg}$, and concatenate relational transformation $\varphi_{rel}$. Given an input vector $v_i^{(j)}$, each layer updates its representation as follows:

\begin{equation}
v_i^{(j)}\leftarrow \varphi_{rel}\left(g_{enc}(v_i^{(j)}), \varphi_{agg}\left(\left[v_k^{(j)}\ |\ \forall\ v_k^{(j)}\in v^{(j)}\right]\right)\right).
\end{equation}

The final polyline representation is derived using another max-pooling aggregation over all vectors. Then, the model uses a cross-attention global graph to learn higher-order interaction between the target agent and other elements in the scenario. Finally, it outputs a latent polyline feature modeling the previous trajectory of the target agent combined with its interactions within the scenario. A simple MLP then uses this feature to decode and predict the future trajectory.

\begin{figure}[thbp]
    \centering
    \includegraphics[width=\linewidth]{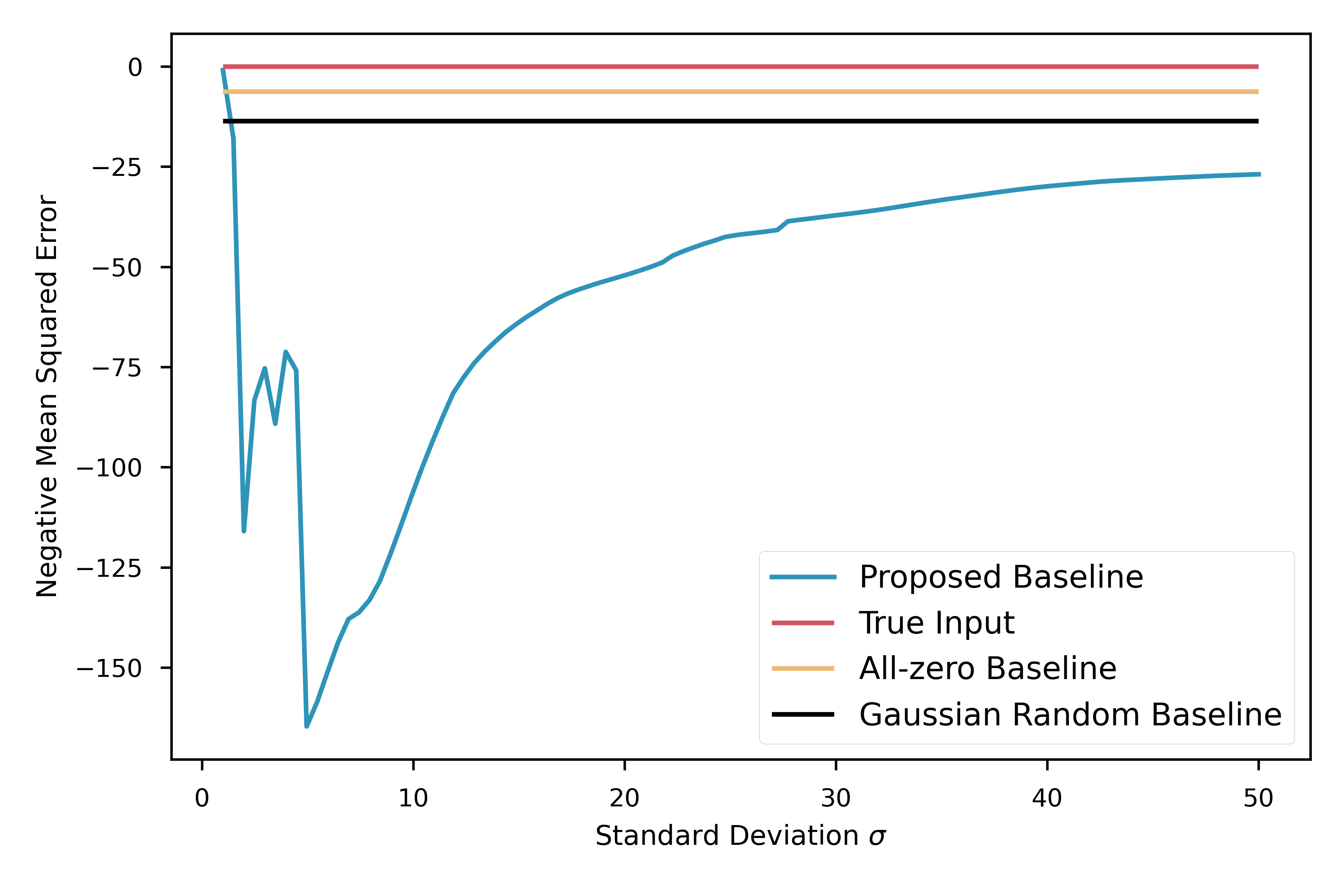}
    \caption{Example visualization of performance differences between the actual input and generated baselines. The proposed baseline (blue curve) generated using different standard deviations has significantly lower accuracy regarding the actual input than the others.}
    \label{fig:baseline}
\end{figure}

\subsection{Generalizability Test and Feature Attribution}

The generalizability of a model refers to its ability to retain performance on data from a shifted domain. Based on this definition, we propose to evaluate a model's generalizability using a cross-scenario validation. Assuming driving behavior is highly similar, a model is first trained and tested on data observed from the same traffic scenario to obtain a "benchmark" performance. Then we further test the model on data from other scenarios. The difference in prediction accuracy compared to the benchmark reflects the generalizability. The procedure simulates an Out-of-Distribution use case where test scenarios are partially or entirely unseen during the training.

Generalizability can be affected by various factors, and spurious correlation between irrelevant features and derived prediction is one of the major causes \cite{cogswell2016reducing}. This paper uses the Integrated Gradients (IG) as the feature attribution to reveal how a model weighs the input features. It helps determine the origin of low generalizability. IG calculates the element-wise product of an input and the gradients with respect to it to derive the "gradient saturation" as an alternative measure. It then aggregates them on an interpolated straight-line path between the input and a baseline, where the counterfactual baseline ideally should output the worst accuracy. Suppose we denote an input feature and its corresponding baseline by $x_i$ and $x_i^\prime$, respectively. In that case, its IG is estimated using the Riemann sum approximation:
\begin{equation}
    \hat{\mathrm{IG}}_i:=(x_i-x_i^\prime)\times\sum_{k=1}^m\frac{\partial F(x_i^\prime+\frac{k}{m}(x_i-x_i^\prime))}{\partial x_i}\times\frac{1}{m},
\end{equation}
where $F$ is the model output, $m$ is the total number of interpolated points (i.e., the number of steps for approximation). In its formulation, definitions of a baseline input $x^\prime$ and prediction score function $F$ are critical for the rationality of this attribution method. The existing works mainly apply this method in classification tasks. They generate baseline inputs with all-zero padding (e.g., a black image), blur kernels (e.g., a Gaussian kernel), or random noises. The output $F$ is the model-predicted probability of the correct class. However, the trajectory prediction task in this paper is a regression instead of a classification problem. Features in the input graph are heterogeneous, and the expected output is not a scalar probability.

In this paper, we address these problems by replacing the predicted correct class's probability in the original implementation with a negative mean squared error (NMSE). The feature attribution to the NMSE reflects a feature's contribution to improving the prediction accuracy. Following the idea that a baseline input should represent the "missing" of features and lead to worst accuracy:
\begin{itemize}
    \item \textbf{All-zero padding for discrete features}. Recall that a vector includes a one-hot encoding $\mathbf{a}^{(j)}$ denoting the specific polyline type in our application. We replace it with a zero vector in the baseline inputs, referring to an invalid polyline type. Similarly, we set polyline id $j=0$ in the baseline, referring to non-existence.
    \item \textbf{Gaussian noise for continuous features}. A zero value could be noteworthy for continuous variables like coordinates, velocities, and headings. For example, a zero velocity means the vehicle stops at the current frame, significantly affecting the possible position in the next frame. Hence, inspired by the work \cite{smilkov2017smoothgrad}, we sample a noise term from a Gaussian distribution and use it to update the original feature.
    \begin{equation}
        x_i\leftarrow x_i+\epsilon_i,\quad\epsilon_i\sim\mathcal{N}\left(\bar{x_i},\sigma\right).
    \end{equation}
\end{itemize}
The original paper pointed out that a counterfactual baseline should correspond to a zero score \cite{pmlr-v70-sundararajan17a}. We propose to evaluate this property of a baseline by comparing prediction accuracy on different input settings. Fig. 2 shows an example with a standard deviation ranging from 0 to 50. The model performs significantly worse on the proposed baseline than the actual input and all-zero baseline. Compared to a baseline with all its elements generated from the same Gaussian distribution, our baseline holds a lower score in most cases. Therefore, it's closer to the actual null baseline the attribution method requires. 

\begin{table*}[t!]
    \centering
    \caption{Generalization Performance of VectorNet}
    \resizebox{\textwidth}{!}{%
    \begin{tabular}{@{}lccccccc@{}}
        \toprule
        \multicolumn{1}{c}{} &
           &
          \multicolumn{6}{c}{\textbf{Training Scenarios}} \\ \cmidrule(l){3-8} 
        \multicolumn{1}{c}{\multirow{-2}{*}{\textbf{Test Scenarios}}} &
          \multirow{-2}{*}{\textbf{Metrics}} &
          \textbf{\begin{tabular}[c]{@{}c@{}}CHN\\ Merging ZS0\end{tabular}} &
          \textbf{\begin{tabular}[c]{@{}c@{}}CHN\\ Merging ZS2\end{tabular}} &
          \textbf{\begin{tabular}[c]{@{}c@{}}DEU\\ Roundabout OF\end{tabular}} &
          \textbf{\begin{tabular}[c]{@{}c@{}}USA\\ Roundabout FT\end{tabular}} &
          \textbf{\begin{tabular}[c]{@{}c@{}}USA\\ Intersection GL\end{tabular}} &
          \textbf{\begin{tabular}[c]{@{}c@{}}USA\\ Intersection MA\end{tabular}} \\ \midrule
         &
          minADE &
          \cellcolor[HTML]{CFF09E}\textbf{0.0193 (0.0003)} &
          0.1154 (0.0004) &
          0.2737 (0.0082) &
          0.0759 (0.0018) &
          0.1376 (0.0040) &
          0.1902 (0.0040) \\
         &
          minFDE &
          \cellcolor[HTML]{CFF09E}\textbf{0.0349 (0.0007)} &
          0.2756 (0.0020) &
          1.0216 (0.0206) &
          0.2325 (0.0017) &
          0.4974 (0.0047) &
          0.6012 (0.0119) \\
        \multirow{-3}{*}{\textbf{CHN Merging ZS0}} &
          MR &
          \cellcolor[HTML]{CFF09E}\textbf{0.0585 (0.0000)} &
          0.7211 (0.0000) &
          0.8430 (0.0000) &
          0.5520 (0.0001) &
          0.7848 (0.0001) &
          0.9133 (0.0000) \\ \midrule
         &
          minADE &
          0.0702 (0.0010) &
          \cellcolor[HTML]{CFF09E}\textbf{0.0209 (0.0001)} &
          0.2656 (0.0070) &
          0.0948 (0.0028) &
          0.2748 (0.0034) &
          0.0546 (0.0041) \\
         &
          minFDE &
          0.1594 (0.0023) &
          \cellcolor[HTML]{CFF09E}\textbf{0.0461 (0.0015)} &
          1.1338 (0.0377) &
          0.1648 (0.0020) &
          0.6427 (0.0065) &
          0.2113 (0.0058) \\
        \multirow{-3}{*}{\textbf{CHN Merging ZS2}} &
          MR &
          0.3697 (0.0006) &
          \cellcolor[HTML]{CFF09E}\textbf{0.0963 (0.0002)} &
          0.5648 (0.0002) &
          0.2823 (0.0003) &
          0.7861 (0.0003) &
          0.3365 (0.0002) \\ \midrule
         &
          minADE &
          0.0771 (0.0336) &
          0.0914 (0.0597) &
          \cellcolor[HTML]{CFF09E}\textbf{0.0277 (0.0121)} &
          0.0188 (0.0061) &
          0.0424 (0.0269) &
          0.0486 (0.0221) \\
         &
          minFDE &
          0.4469 (0.0407) &
          0.5773 (0.1093) &
          \cellcolor[HTML]{CFF09E}\textbf{0.1691 (0.0349)} &
          0.1482 (0.0245) &
          0.2479 (0.0642) &
          0.3367 (0.0612) \\
        \multirow{-3}{*}{\textbf{DEU Roundabout OF}} &
          MR &
          0.7332 (0.0086) &
          0.8898 (0.0100) &
          \cellcolor[HTML]{CFF09E}\textbf{0.4553 (0.0050)} &
          0.4283 (0.0000) &
          0.3985 (0.0050) &
          0.5715 (0.0086) \\ \midrule
         &
          minADE &
          0.1534 (0.0066) &
          0.1201 (0.0063) &
          0.1359 (0.0054) &
          \cellcolor[HTML]{CFF09E}\textbf{0.0186 (0.0017)} &
          0.0483 (0.0003) &
          0.0499 (0.0026) \\
         &
          minFDE &
          0.6942 (0.0147) &
          0.5415 (0.0071) &
          0.5763 (0.0025) &
          \cellcolor[HTML]{CFF09E}\textbf{0.0866 (0.0010)} &
          0.1736 (0.0026) &
          0.2489 (0.0026) \\
        \multirow{-3}{*}{\textbf{USA Roundabout FT}} &
          MR &
          0.6470 (0.0000) &
          0.7199 (0.0000) &
          0.3667 (0.0000) &
          \cellcolor[HTML]{CFF09E}\textbf{0.2074 (0.0000)} &
          0.6215 (0.0000) &
          0.3421 (0.0000) \\ \midrule
         &
          minADE &
          0.1117 (0.0052) &
          0.0783 (0.0005) &
          0.0671 (0.0010) &
          0.0246 (0.0003) &
          \cellcolor[HTML]{CFF09E}\textbf{0.0120 (0.0003)} &
          0.0520 (0.0014) \\
         &
          minFDE &
          0.6877 (0.0083) &
          0.5670 (0.0115) &
          0.3601 (0.0017) &
          0.1422 (0.0011) &
          \cellcolor[HTML]{CFF09E}\textbf{0.0566 (0.0004)} &
          0.2511 (0.0025) \\
        \multirow{-3}{*}{\textbf{USA Intersection GL}} &
          MR &
          0.8898 (0.0000) &
          0.8944 (0.0000) &
          0.7515 (0.0001) &
          0.5281 (0.0001) &
          \cellcolor[HTML]{CFF09E}\textbf{0.2748 (0.0001)} &
          0.6426 (0.0001) \\ \midrule
         &
          minADE &
          0.1212 (0.0078) &
          0.1324 (0.0066) &
          0.1030 (0.0048) &
          0.0483 (0.0009) &
          0.0718 (0.0023) &
          \cellcolor[HTML]{CFF09E}\textbf{0.0261 (0.0005)} \\
         &
          minFDE &
          0.4263 (0.0040) &
          0.4283 (0.0100) &
          0.3562 (0.0071) &
          0.2051 (0.0047) &
          0.2297 (0.0035) &
          \cellcolor[HTML]{CFF09E}\textbf{0.0934 (0.0018)} \\
        \multirow{-3}{*}{\textbf{USA Intersection MA}} &
          MR &
          0.6030 (0.0004) &
          0.7027 (0.0004) &
          0.3783 (0.0004) &
          0.3872 (0.0001) &
          0.4164 (0.0005) &
          \cellcolor[HTML]{CFF09E}\textbf{0.2045 (0.0004)} \\ \bottomrule
    \end{tabular}
    }
    \label{tab:2}
\end{table*}

\section{EXPERIMENT RESULTS}
\subsection{Experimental Setup}

In this paper, we use the INTERACTION dataset \cite{interactiondataset} to test the generalization performance of the VectorNet. Specifically, we select six different scenarios of three categories: expressway merging, unsignalized intersections, and roundabouts. Table I shows the details of the dataset for training and testing. The task is to predict the future 3-second trajectory of a single target agent based on the last 1-second observations. We train our model with a batch size of 64 and an initial learning rate of 0.001. The learning rate decays to 30\% every five epochs. Besides, our training process excludes the auxiliary node completion task for that it only improves the performance by four to six percent \cite{Gao_2020_CVPR}.

We evaluate our model using minimum Average Displacement Error (minADE), minimum Final Displacement Error (minFDE), and Miss Rate (MR). We use MR as the critical metric to determine the performance difference. Specifically, for any vehicle in a case, if its prediction at the final frame is out of a lateral or longitudinal threshold with respect to the ground truth, it's considered a "miss" case, and MR is the percentages of miss cases in a batch. The lateral threshold is 1 meter. Calculations of the first two metrics and longitudinal threshold are as follows:

\begin{align}
    \text{minADE} &= \min_n\frac{1}{T}\sum_{t}\sqrt{(\hat{x}_t-x_t)^2+(\hat{y}_t-y_t)^2}, \\
    \text{minFDE} &= \min_n\sqrt{(\hat{x}_T-x_T)^2+(\hat{y}_T-y_T)^2,} \\
    \text{Threshold}_{\text{lon}} &= \begin{cases}1 & v<1.4\ \text{m/s} \\ 1 + \frac{v-1.4}{11-1.4} & \text{1.4 m/s}\leq v\leq\text{11 m/s} \\ 2 & v\geq\text{11 m/s}\end{cases},
\end{align}
where $n$ is the sample index in a batch, $T$ is the prediction horizon in frames, and $v$ is the ground truth target agent speed at the last frame. All calculations have the same unit of meters.

\begin{figure*}[t!]
    \centering
    \includegraphics[width=6.8in]{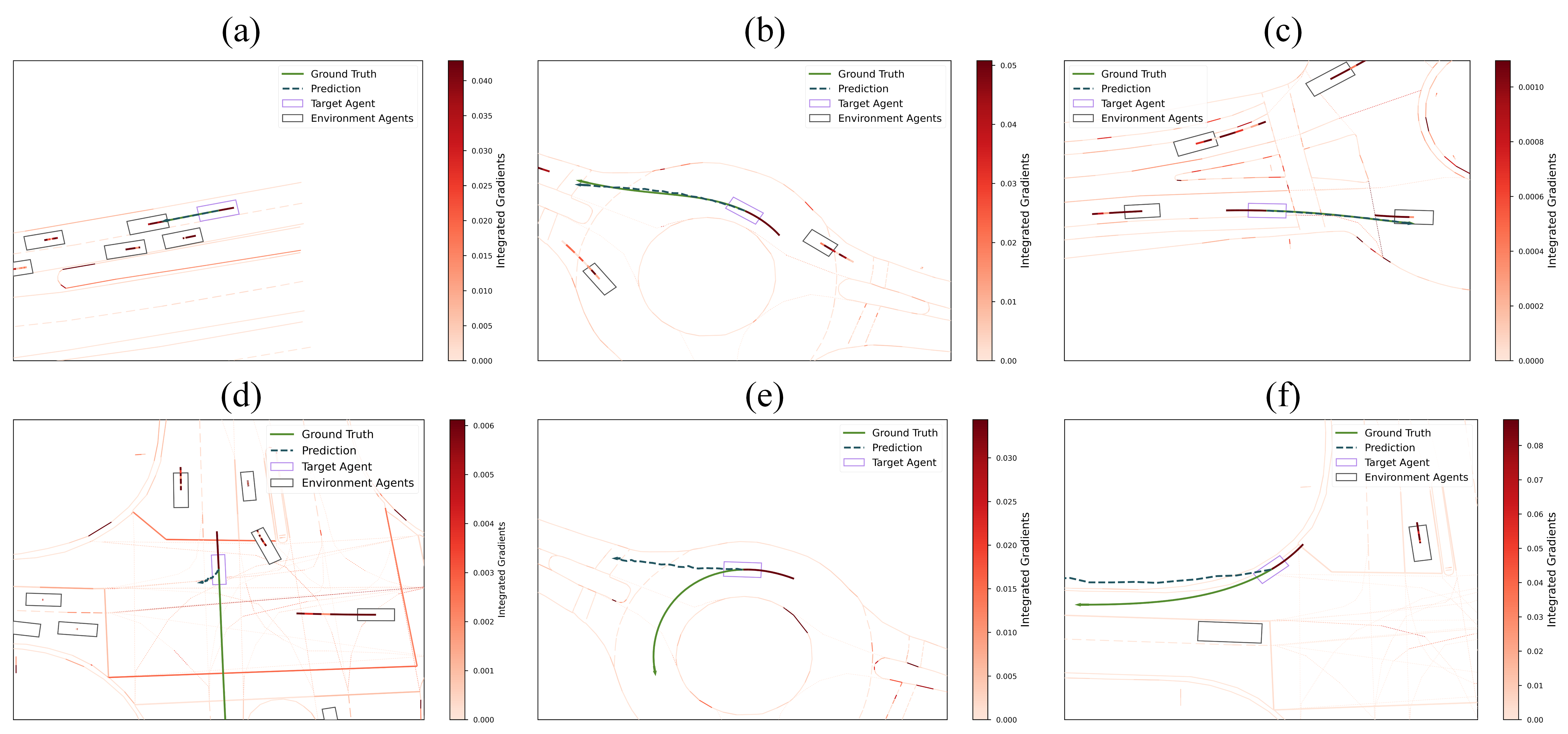}
    \caption{Selected best-case (top row) and typical worst-case prediction (bottom row) results. Input vectors (i.e. map polylines and history trajectories) are colored by their absolute values of the sum integrated gradients over the feature dimension. A darker coloring indicates a higher relevance to the final prediction. Training and testing settings are: (a) model trained on CHN Merging ZS0 and tested on the same scenario; (b) model trained on CHN Merging ZS0 and tested on USA Roundabout FT; (c) model trained on USA Intersection MA and tested on DEU Roundabout OF; (d) model trained on CHN Merging ZS2 and tested on USA Intersection MA; (e) model trained on DEU Roundabout OF and tested on the same scenario; (f) model trained on USA Intersection GL and tested on USA Intersection MA.}
    \label{fig:qualitative}
\end{figure*}

\subsection{Generalizability Test}

The validation results of models trained on different data are tabulated in Table II. The diagonal values demonstrate a model's "baseline" performance validated using observations from the exact scenario. Based on the MR, the VectorNet achieves the highest score of 5.85\% on the CHN Merging ZS0 and the lowest 46.23\% on the DEU Roundabout OF. The reason for this is two-folded. First, the latter has a higher traffic mix, with 5.8\% and 4.1\% non-vehicle ratios in the training and testing dataset. Another cause is the significantly smaller training dataset could potentially lead to insufficient training. In terms of minimum ADE and FDE, results from different test scenarios are close because there are short predictions (i.e., actual trajectory with less than 3 seconds) in all the cases. Such results also demonstrate VectorNet, like many existing models, can perform well under short-term predictions.

In cross-scenario tests, some models demonstrate a "familiarity" property that they maintain the accuracy in similar scenarios but fail to handle new ones. For example, the model trained on CHN Merging ZS0 has the lowest MR increase on CHN Merging ZS2. The two cases have the same base map, but vehicles are at different positions (i.e., on-ramp vs. mainline). And the model trained on DEU Roundabout OF can achieve a better score when running on USA Roundabout FT. Notice the minimum ADE and FDE increase only by around 0.08 and 0.38 meters on average, which means the VectorNet can maintain its performance on short-term prediction. However, drastic increases in the MR in most test cases reveal that the model may fail to learn a generalizable pattern and overfit the data when training on limited observations. The model trained on CHN Merging ZS2 holds the worst generalization performance with the highest MR increase of around 70\%. And one that trained on USA Roundabout FT is overall the best with an averaged miss rate of 40\% across all scenarios. In short, the results reflect that VectorNet has limited generalizability and suffers from severe accuracy bottlenecks on OOD data. To further investigate the potential causes of this problem, we apply the proposed integrated gradient method in the following section to see how the model understands its surroundings by attributing prediction scores to input polyline features.

\subsection{Qualitative Analysis}

Referring to the previous exercises \cite{sayres2019using, Warrick_2018} of visualizing attributions, we color the input vectors associated with map polylines and history trajectories by their absolute values of the sum integrated gradients over the feature dimension. A darker coloring indicates a more important vector for the final prediction. To avoid domination of the color scheme by extreme attribution values, we normalize the shading using the 99th percentile.

The top row of Fig. 3 shows three selected best prediction results of our model. It can predict a future trajectory with high precision when trained and tested on the same scenario, as shown in Fig. 3(a). When the testing scenario differs from the training scenario, the model is also able to have great prediction accuracy both when there exists a leading vehicle (see Fig. 3(c)) or not (see Fig. 3(b)). In addition to showing the good-performed cases, we also identify three typical cases where the model usually fails and plot their representatives in the next row. Below are detailed discussions of these issues:
\begin{itemize}
    \item \textbf{Ineffectiveness}. This issue appears when the model is tested on a more complicated scenario than the one it was trained on. As in Fig. 3(d), VectorNet trained on CHN Merging ZS2 is unable to utilize virtual lines and stop lines in the intersection scenario to guide their estimation for future trajectories, even though having assigned these vectors with relatively high attributions. This is due to the fact that these types of information are OOD features for the training dataset. The model conceives no perceptions of their correct relationship to the proper trajectory and assigns ineffective relevance.
    \item \textbf{Blindfold}. This problem commonly exists in all the scenarios when models are misled when assigning excessive attribution to a single or a few polylines. In Fig 3(b), the model assigns high attributions to the border of the roundabout to the right side of the driving direction, which promotes a correct prediction on exiting the roundabout. On the other hand, the model shown in Fig 3(e) assigns its historical trajectory with a significantly higher attribution than the rest of the polylines. When the prediction is heavily dependent on the previous trajectory, the model fails to realize the possibility that a vehicle is turning left within the roundabout.
    \item \textbf{Disregard}. Another critical cause for performance deterioration is disregard. Unlike the aforementioned ineffectiveness, disregard refers to the model assigning important polylines with low relevance to the prediction though having been trained on data with these elements. In Fig. 3(f), we see a model trained on USA Intersection GL fails to address the road border to the right while turning in USA Intersection MA. Instead, it assigns high importance to an irrelevant vehicle at the upper right. As a result, though it well estimates the intention, the predicted trajectory is irrational since it disregards the essential constraint to keep on the vehicle lanes.
\end{itemize}


\subsection{Quantitative Analysis}

This section gives a quantitative analysis of the model by showing which vectors have higher relevance to the final prediction. We narrow our focus on the best and worst models judged by their average MR in Table II. They are the ones trained on USA Roundabout FT and CHN Merging ZS2, respectively. We test both of them on the same Intersection GL scenario to highlight differences between the training and testing datasets and maintain comparability.


\begin{figure}[thbp]
    \centering
    \includegraphics[width=\linewidth]{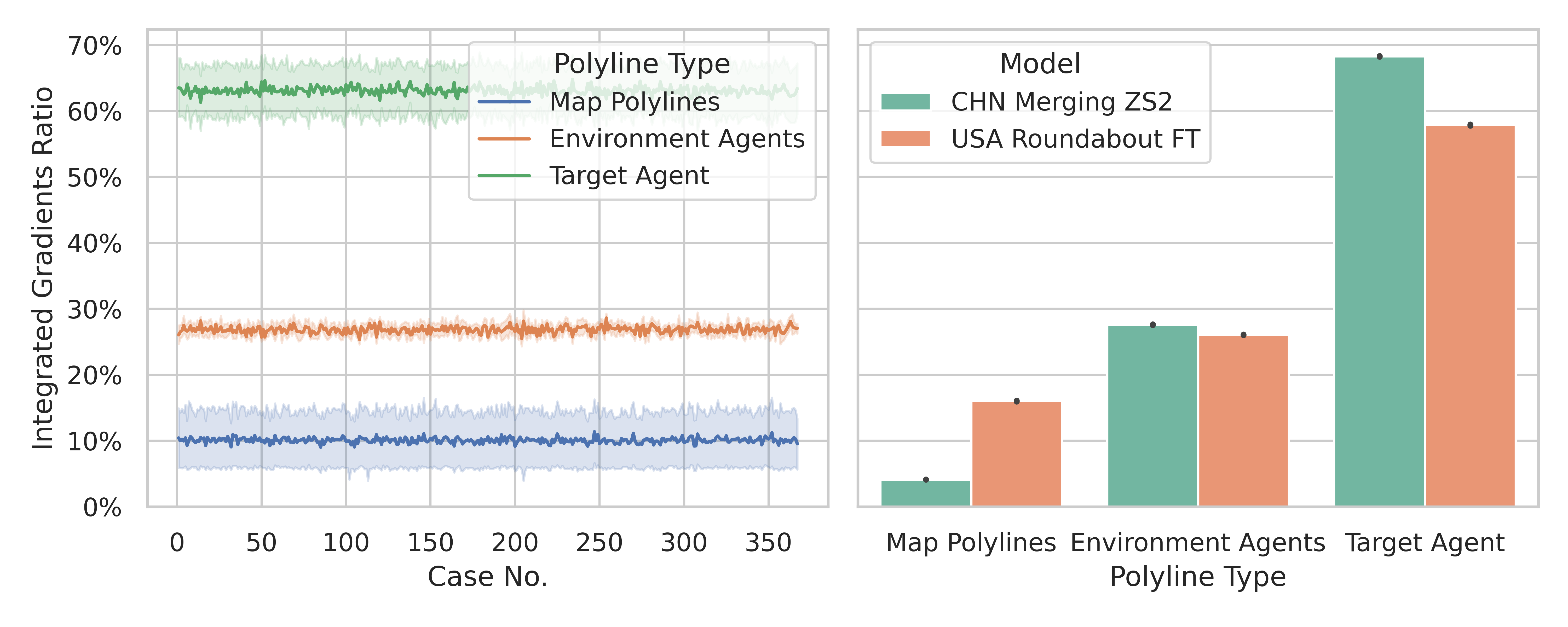}
    \caption{Average feature attribution ratios aggregated by polyline types with respect to different test cases in USA Intersection GL and total feature attribution (right) of models trained on CHN Merging ZS2 and USA Roundabout FT.
    }
    \label{fig:quantative_pline}
\end{figure}

\emph{1) Analysis of polyline weighting:} The visualization uses the sum of integrated gradients to help demonstrate the relative importance among different types of polylines. In Fig. 4, the result shows a consistent tendency of weighing strategy for the two models across all the cases. History trajectories of the target agents rank the most important vectors in the scenes. The ratios of target agent trajectory, environment agent trajectory, and map polylines are 68:28:4 and 58:26:16 for the two models, respectively. The meager relevance assigned to map polylines by the first model explains its low prediction accuracy caused by ignorance or disregard. Trained on a simple expressway mainline scenario of CHN Merging ZS2, the model learns to follow the movements of the surrounding vehicles (corresponding to a high ratio of importance for trajectories) and predict its future trends mainly based on the history track. The second model has a more balanced weight assignment with increased attention on the map polylines. We believe more diverse and intensive interactions in the roundabout scenario contribute to this difference. In short, this analysis reveals the drawback of training a trajectory prediction model on a limited dataset. Especially when the case is relatively more straightforward, the model can easily overfit a specific type of input, deteriorating the generalizability.

\emph{2) Analysis of feature weighting:} Section III mentioned five features used to construct the vectors, including the origin and destination coordinates, type encoding, motion status, and polyline id. How the model determines the importance of these features also affects the outcome performance. To illustrate the idea, suppose driving in an intersection, we would expect the driver to pay attention not only to surrounding moving vehicles but also nearby infrastructures. A skewed attribution distribution for different features can lead to wrong judgment and irrational actions. Applying this idea to our model analysis, we find that the two models both weight motion status as the most important feature, followed by origin and destination coordinates (see Fig. 5). It shows that models pay substantially more attention to moving objects, which is consistent with the results in section IV.D.1). The massive attribution gap between the motion feature and other features indicates that the former has dominated the prediction, pushing the models to disregard context-related constraints. However, the model trained on USA Roundabout FT has a more balanced relevance distribution with increased attention to polyline types compared to the one trained on CHN Merging ZS0. Such balanced relevance distribution can potentially be an important factor for better generalizability of the model, as seen in Table II.

\begin{figure}[thbp]
    \centering
    \includegraphics[width=\linewidth]{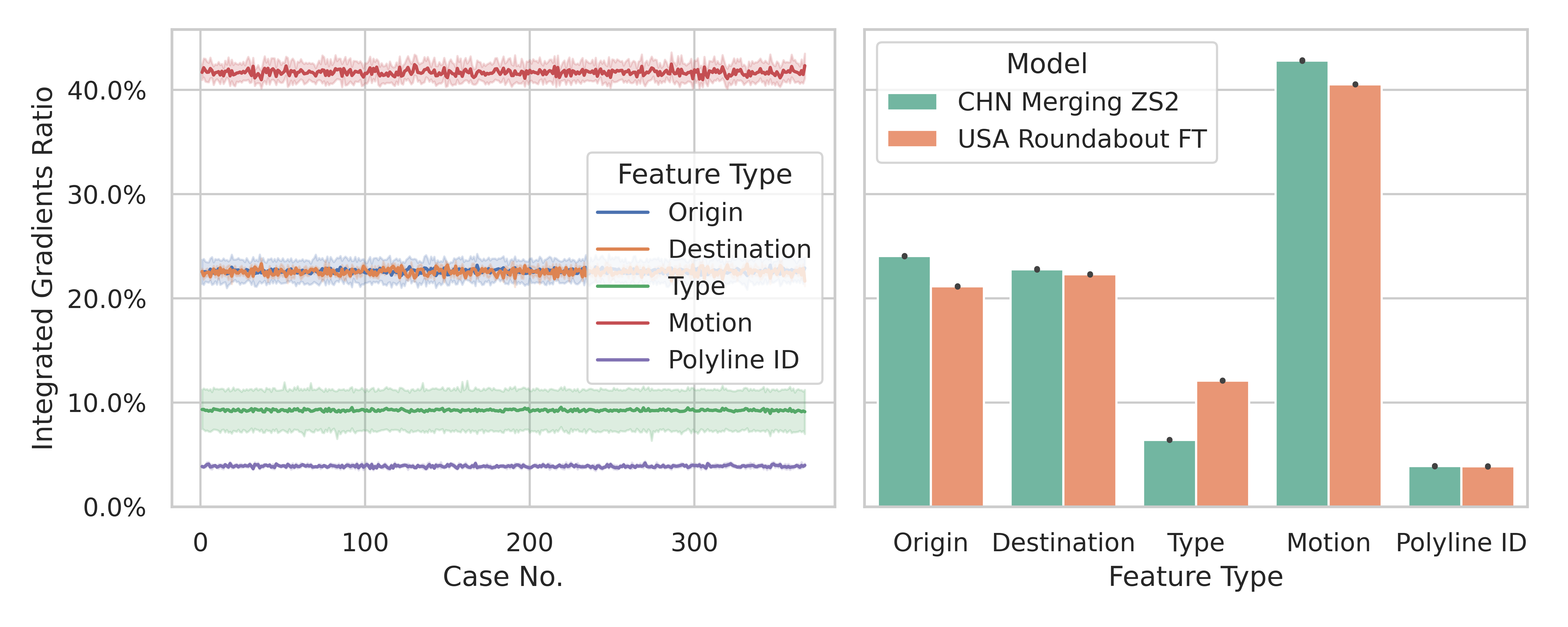}
    \caption{Average feature attribution ratios aggregated by node features types with respect to different test cases in USA Intersection GL and total feature attribution (right) of models trained on CHN Merging ZS2 and USA Roundabout FT.
    }
    \label{fig:quantative_feature}
\end{figure}

\section{CONCLUSION}

This paper analyzes the generalizability of a state-of-the-art graph-based trajectory predictor with vectorized representation, VectorNet. To imitate the OOD setting, we only select data from one scenario for training at each time and use data from the remaining scenarios for testing. The training and test scenarios are usually different in both road topology and traffic rules. The experiment result of our study shows a significant increase in both displacement errors and miss rate, corresponding to the low generalizability of the model. Integrated gradients are used in qualitative and quantitative analysis to understand the cause. We identify three typical issues that lead to generalization bottlenecks: \textit{Ineffectiveness}, \textit{Blindfold}, and \textit{Disregard}, and link them to an imbalanced weighting of polylines and features. One limitation of the proposed attribution method is that it can only reflect if the model properly assigns a high relevance to important features related to correct prediction. It fails to explain how the model deduces causal relationships from these correlations, which will be further discovered in our future works. In general, we hope this paper can promote research to address this limitation and investigate generalizability for a broader range of trajectory prediction models.

\section*{ACKNOWLEDGMENT}

We would like to thank Jie Feng for help on the initial model implementation and Wei-Jer Chang for insightful discussions.




\addtolength{\textheight}{-0cm}   








\bibliographystyle{IEEEtran}
\bibliography{IEEEabrv}

\end{document}